\documentclass[letterpaper, 10 pt, conference]{ieeeconf}  

\IEEEoverridecommandlockouts   

\usepackage{graphics} 
\usepackage{epsfig} 
\usepackage{mathptmx} 
\usepackage{times} 
\usepackage{amsmath} 
\usepackage{amssymb}  
\usepackage{booktabs}
\usepackage{tabularx}
\usepackage{makecell}
\usepackage{caption}
\usepackage{subcaption}  

\usepackage{cuted}      
\usepackage{graphicx}   
\usepackage{caption}    

\usepackage[dvipsnames,table,xcdraw, HTML]{xcolor}
\usepackage{hyperref}
\hypersetup{
    colorlinks=true,
    linkcolor=orange,
    filecolor=magenta,      
    urlcolor=orange,
    citecolor=orange,
}
\usepackage[capitalise, nameinlink]{cleveref}
\makeatletter
\let\NAT@parse\undefined
\makeatother
\usepackage[numbers,sort,compress]{natbib}
\usepackage{caption}
\newcommand{\RR}{\mathbb{R}}

\newcommand{\set}[1]{\left\lbrace #1 \right\rbrace}

\title{\LARGE \bf
\textsc{Motion Tracks}: A Unified Representation for \\Human-Robot Transfer in Few-Shot Imitation Learning
}

\author{Juntao Ren\textsuperscript{1}, %
Priya Sundaresan\textsuperscript{2}, %
Dorsa Sadigh\textsuperscript{2}, %
Sanjiban Choudhury\textsuperscript{1}, %
Jeannette Bohg\textsuperscript{2}}

\begin{document}

\twocolumn[{
\renewcommand\twocolumn[1][]{#1}
\maketitle
\begin{center}
\captionsetup{type=figure}
\vspace{-0.5cm}
\includegraphics[width=0.88\textwidth]{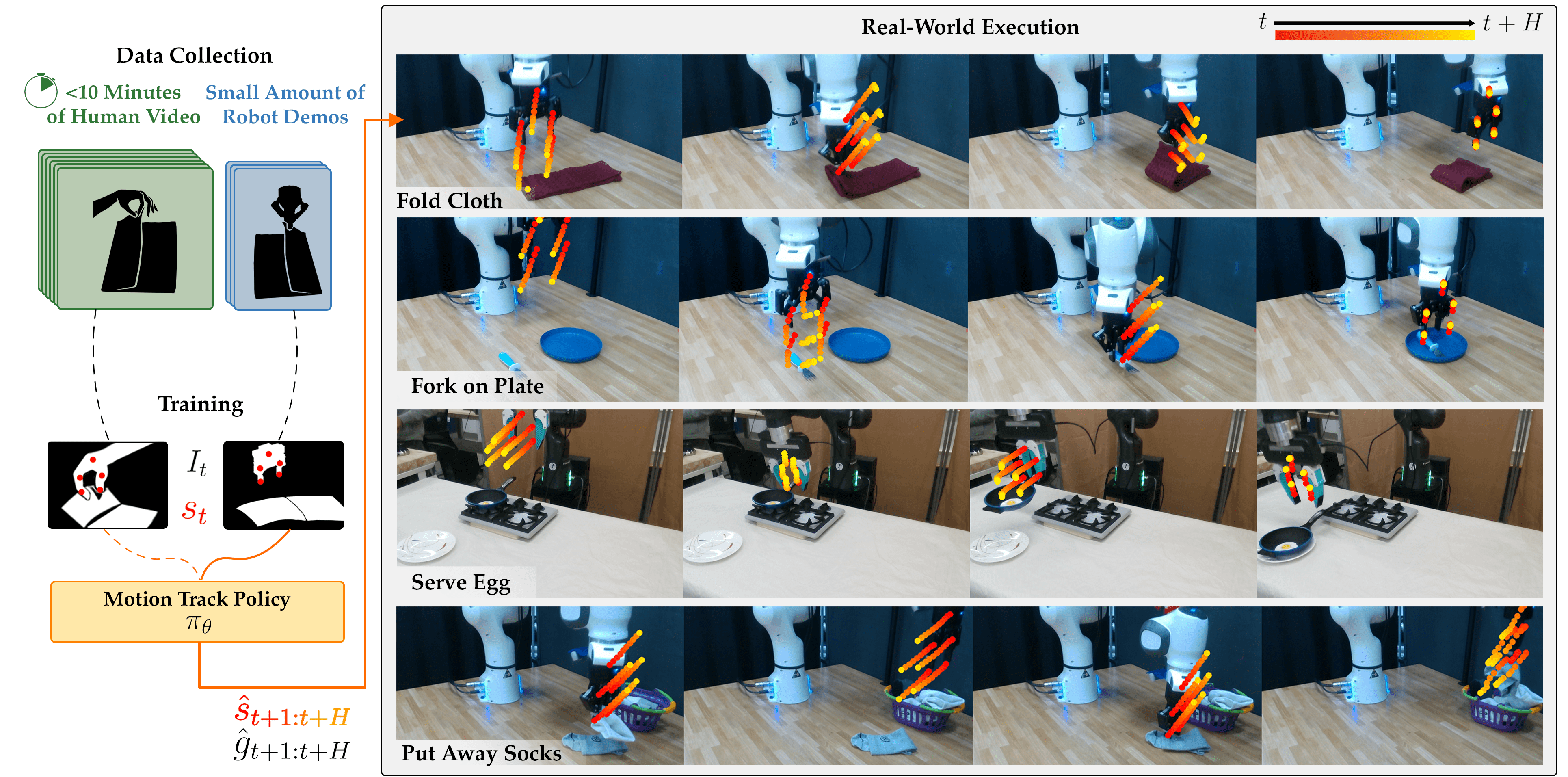}
\captionof{figure}{\textbf{Motion Track Policy (MT-$\pi$ Overview)}: Left: Using 10 minutes of human video and a small set of robot demonstrations, we train MT-$\pi$ to take a third-person image observation ($I_t$) and output \emph{motion tracks} ($\hat{s}$), a cross-embodiment action space of 2D trajectories in image space representing manipulator movement. Right: At test-time, we predict $\hat{s}$ along with grasps ($\hat{g}$) from 2 camera views (1 shown) and recover full 6DoF robot actions using stereo triangulation for execution.}\label{fig:front}
\end{center}
}]

\thispagestyle{empty}
\pagestyle{empty}

\renewcommand{\thefootnote}{}
\footnote{\textsuperscript{1}Cornell University \textsuperscript{2}Stanford University. Correspondence to {\tt jlr429@cornell.edu}}


\begin{abstract}
Teaching robots to autonomously complete everyday tasks remains a challenge. Imitation Learning (IL) is a powerful approach that imbues robots with skills via demonstrations, but is limited by the labor-intensive process of collecting teleoperated robot data. Human videos offer a scalable alternative, but it remains difficult to directly train IL policies from them due to the lack of robot action labels. To address this, we propose to represent actions as short-horizon 2D trajectories on an image. These actions, or \emph{motion tracks}, capture the predicted direction of motion for both human hands and robot end-effectors. We instantiate an IL policy called Motion Track Policy (MT-$\pi$) which receives image observations and outputs motion tracks as actions. By leveraging this unified, cross-embodiment action space, \mbox{MT-$\pi$} completes tasks with high success given just minutes of human video and limited additional robot demonstrations. At test time, we predict motion tracks from two camera views, recovering 6DoF trajectories via multi-view synthesis. \mbox{MT-$\pi$} achieves an average success rate of 86.5\% across 4 real-world tasks, outperforming state-of-the-art IL baselines which do not leverage human data or our action space by 40\%, and generalizes to scenarios seen only in human videos. Code and videos are available on our website \href{https://portal-cornell.github.io/motion_track_policy/}{(https://portal-cornell.github.io/motion\_track\_policy/)}.
\end{abstract}
\section{Introduction}

Imitation learning (IL) is a widely adopted approach for training robot policies from human demonstrations~\cite{schaal1996learning, hussein2017imitation}. However, even state-of-the-art IL policies can in some cases require on the order of hundreds~\cite{chi2023diffusion, zhao2023learning} or up to \emph{tens of thousands}~\cite{padalkar2023open, team2024octo, khazatsky2024droid} of crowdsourced, teleoperated demonstrations in order to achieve decent performance.  These demonstrations are typically collected by teleoperating robots via virtual reality devices~\cite{OrbikEbert2021OculusReader} or puppeteering interfaces~\cite{zhao2023learning, fu2024mobile, wu2023gello}. However, not only are some devices robot-specific and not universally accessible, prolonged teleoperation with them is time-consuming, labor-intensive, and often requires extensive practice before dataset collection can even begin.

An alternative approach is robot learning from passive observations, such as videos of humans demonstrating desired tasks. Human video data is far easier to collect at scale, with existing datasets providing thousands of hours of demonstrations ~\cite{grauman2022ego4d, goyal2017something}. However, these human videos lack the robot action labels that are necessary for training imitation learning policies, presenting a challenge in transferring knowledge from human videos to robot policies.

Recent works have addressed this problem by pretraining visual representations on human data~\cite{nair2022r3m, karamcheti2023language, grauman2022ego4d}. However, these representations often prove challenging to apply directly or fine-tune for specific tasks, due to the diversity of pretraining data~\cite{xie2024decomposing, dasari2023unbiased}. Alternative approaches focus on collecting task-specific human videos and teleoperated robot demonstrations, which capture a broader range of motions. These approaches then learn shared state space representations~\cite{wen2023any, bharadhwaj2024track2act, jain2024vid2robot}, shared latent embeddings~\cite{wang2023mimicplay}, or reward functions through inverse reinforcement learning~\cite{zakka2022xirl}. While these methods align input representations effectively, their reliance on robot datasets for output actions limits their expressivity. For example, a robot trained only on rightward drawer-closing demonstrations may fail to generalize to leftward drawer-closing, regardless of the conditioning quality.

Our key insight is that, despite the embodiment gap between human hands and robot end-effectors, we can unify their action spaces by projecting their movements onto the image plane as 2D trajectories. We train an IL policy that takes image observations as input and predicts actions as \emph{motion tracks}, or short-horizon 2D trajectories in image space indicating the predicted direction of motion for points on a human hand or robot end-effector. This reframing makes action prediction compatible with both human hands and robot end-effectors. Our approach requires minimal data: approximately 10 minutes of human video and a few dozen robot demonstrations. We obtain ground-truth motion tracks using state-of-the-art hand-tracking~\cite{pavlakos2024reconstructing} for human observations and forward kinematics with known camera extrinsics for robot demonstrations. At test time, we predict motion tracks from two camera views and use multi-view geometry to reconstruct 6DoF end-effector trajectories in 3D.

In short, we present Motion Track Policy (MT-$\pi$), which
\begin{enumerate}
    \item Is a simple IL policy trainable from \textbf{minutes} of easy-to-collect human video data and a modest amount of additional teleoperated robot data,
    \item Achieves an average 86.5\% success rate across 4 real-world tasks, which is 40\% higher compared to SOTA IL approaches that do not make use of human video or our action space,
    \item Demonstrates generalization to novel scenarios only captured in human videos, and
    \item Is open-sourced for reproducible training and deployment in the real world.
\end{enumerate}

\section{Related Work}

\subsection{Imitation Learning}
Imitation learning (IL) is a framework where an agent learns to mimic expert behavior by observing demonstrations~\cite{pomerleau1988alvinn, schaal1996learning, atkeson1997robot}. This paradigm has been widely applied in robotics, where policies are trained to map sensory inputs to motor actions~\cite{levine2016end, torabi2018behavioral, duan2017one}. Recent advancements in policy architectures, such as Diffusion Policy~\cite{chi2023diffusion} and Action Chunking Transformer (ACT)~\cite{zhao2023learning}, have significantly improved the real-world applicability of IL. However, these approaches still often require a significant amount of teleoperated data to achieve decent performance.

Several works have aimed to streamline teleoperation with easy-to-use interfaces like puppeteering~\cite{zhao2023learning, wu2023gello}, mirroring of tracked human hands~\cite{ding2024bunny, shaw2024learning, park2024using}, or intuitive hand-held devices~\cite{chi2024universal}. These setups can make data collection more convenient to scale. Large-scale crowdsourcing efforts have recently brought some open-source robotics datasets to the order of more than \emph{80K trajectories}~\cite{khazatsky2024droid, stone2023open}. Nonetheless, in many practical scenarios, collecting data at this scale is infeasible, and learning performant policies from such diverse data remains an open challenge~\cite{team2024octo, kim2024openvla}. This drives the need for training policies in the low-robot-data regime.

\subsection{Sample-Efficient State-Action Representations}
To address the issue of sample inefficiency in IL, recent works propose leveraging alternative input modalities beyond images, such as point clouds~\cite{sundaresan2023kite, ze20243d, goyal2023rvt}, radiance fields~\cite{wang2023sparsedff, rashid2023language}, or voxel grids~\cite{shridhar2023perceiver, yuan2024general}. These representations better capture 3D geometric features in visual scenes, which can lead to decent policies from fewer demonstrations. However, these approaches are only compatible with robot-only data consisting of RGB-D observations, along with calibrated camera extrinsics. Consequently, images remain a prevalent choice for policy inputs as they allow for fewer assumptions during data collection and thus are more suitable to be used with cross-embodiment data like human video. 

One effective strategy for methods that take images as input is to additionally condition on 2D representations like optical flow~\cite{lin2024flowretrieval, yuan2024general, ko2023learning}, object segmentation masks~\cite{duan2024manipulate, bharadhwaj2024towards}, bounding boxes~\cite{stone2023open}, or pixelwise object tracks~\cite{xu2024flow, wen2023any, bharadhwaj2024track2act}. While these methods focus on reparameterizing the inputs to IL policies, the action space remains rooted in robot proprioceptive states. This formulation limits the policy to only learning actions present in the teleoperated data. We instead propose an embodiment-agnostic, image-based action space which encourages behaviors captured in robot data as well as those that may only be present in human videos.


\subsection{Learning from Human Video}

\textbf{Pretrained Representations on Internet-Scale Data.} 
Given the challenges associated with collecting teleoperated robot data, human video is a popular alternative source of data that is easier to collect and more widely available. Several approaches use large-scale human video datasets ~\cite{grauman2022ego4d, goyal2017something, andriluka20142d, damen2018scaling} to pretrain visual representations for robot policy learning~\cite{karamcheti2023language, nair2022r3m, baker2022video}. However, prior works have found these pretrained representations to be brittle when used out-of-the-box or finetuned for specific tasks due to the sheer diversity of the datasets they were trained on~\cite{xie2024decomposing, dasari2023unbiased}.

\textbf{Implicit Priors from Human Video} An alternative approach involves collecting task-specific human video data, along with a smaller amount of teleoperated robot data, and jointly learning on the hybrid dataset~\cite{wang2023mimicplay, zakka2022xirl, xu2023xskill}. The primary challenge with this approach is overcoming the embodiment gap, as human video does not provide robot action labels. To address this, some methods attempt to extract priors from human video that can guide robot learning, such as shared human-robot latent embeddings with which to condition robot policies~\cite{wang2023mimicplay,sharma2019third} or reward signals for reinforcement learning (RL)~\cite{zakka2022xirl,ma2022vip,smith2019avid, jonnavittula2024view, mahesheka2024language}. However, the former methods often require a goal image, which is not always available, and the latter demand significant on-policy robot interaction to train the RL agent.

\textbf{Explicit Tracking of Human Hands.} Recent advancements in human-hand tracking~\cite{pavlakos2024reconstructing, goel2023humans, pavlakos2022human,  lugaresi2019mediapipe} demonstrate that we can now  more reliably track the \emph{explicit} motion of human hands in video, instead of only extracting \emph{implicit} priors. Building off of this, several recent works demonstrate few-shot visual imitation from human video alone, without requiring any robot data~\cite{yang2024equivact, yang2024equibot, papagiannis2024r, zhu2024vision, vecerik2024robotap, wu2024learning}. These methods largely assume that detected human hand poses can be directly retargeted to robot end-effectors. However, the significant morphological differences (i.e. differences in size and shape between human and robot hands) can mean that even just replaying a tracked human hand trajectory on a robot end-effector may fail to produce the desired behavior in many cases. Our work builds on these approaches by introducing a unified action space compatible with both human and robot embodiments, where actions are represented as 2D trajectories in the image plane. As the actions themselves are now aligned in a shared space, we disentangle motion differences from visual differences, allowing us to better bridge the embodiment gap and train on a hybrid of human videos and teleoperated robot data.

\section{Problem Formulation}

We aim to learn a visuomotor policy trained on mostly human video and a small amount of robot demonstrations. To do so, we extract actions from a shared representation of pixel-level keypoints and train a policy that outputs actions directly in image space, making the pipeline compatible with either embodiment.

We assume access to a joint, single-task dataset $D = D_{\text{human}} \cup D_{\text{robot}}$ where we have access to $\geq$ 1 camera during data collection, and 2 cameras during test-time. During data collection, we do not make specific assumptions on the viewpoints, but at test-time we assume access to known extrinsics for two of the cameras. The majority of the dataset consists of easy-to-collect human demonstrations $D_{\text{human}}$, while a \textit{small} set is of teleoperated robot demonstrations $D_{\text{robot}}$. While we do \emph{not} assume human and robot demonstrations to be \emph{paired} (i.e., starting from identical initial object states), we do assume that both embodiments perform similar motions given similar states. During data collection, each demonstration is represented by $\{(I_t^{(i)}, s_t^{(i)}, g_t^{(i)})\}_{t=1}^N,$ where $t$ indexes time and $i$ indexes the viewpoint (i.e., Camera 1/2). $I_t^{(i)}$ represents the RGB image captured by camera $i$ at time $t$, and $s_t^{(i)} = \{(u_j^{(i)}, v_j^{(i)})\}_{j=1}^k$ represents the pixels of $k$ keypoints on the end-effector (human hand or robot gripper) in the image $I_t^{(i)}$. For prehensile tasks, $g_t^{(i)} \in \set{0, 1}$ indicates whether a (human/robot) grasp occurs at $t$.

Given an image $I_t^{(i)}$ and the corresponding $k$ end-effector keypoints $s_t^{(i)}$, our objective is to learn a keypoint-conditioned motion track network
\begin{equation}
    (\hat{s}_{t+1:t+H}^{(i)}, \hat{g}_{t+1:t+H}^{(i)}) \sim \pi_\theta(\cdot \mid I_t^{(i)}, s_t^{(i)}).
\end{equation}
Formally, we define $\hat{s}_{t+1:t+H}^{(i)}$ to be the predicted \textit{motion tracks}, which are 2D trajectories in image space that forecast the future pixel locations of the keypoints on the end-effector over a horizon $H$. $\hat{g}_{t+1:t+H}^{(i)}$ represents the corresponding grasp indicators at each future timestep. The final output of our policy lies in $\RR^{(2k + 1) \times H}$. These short-horizon motion tracks are later triangulated to recover 6DoF actions $a_{t:t+H}$ to be executed on the robot (detailed in \cref{sec:action_inference}).
\section{Approach}
\begin{figure}
    \centering
    \includegraphics[width=\columnwidth]{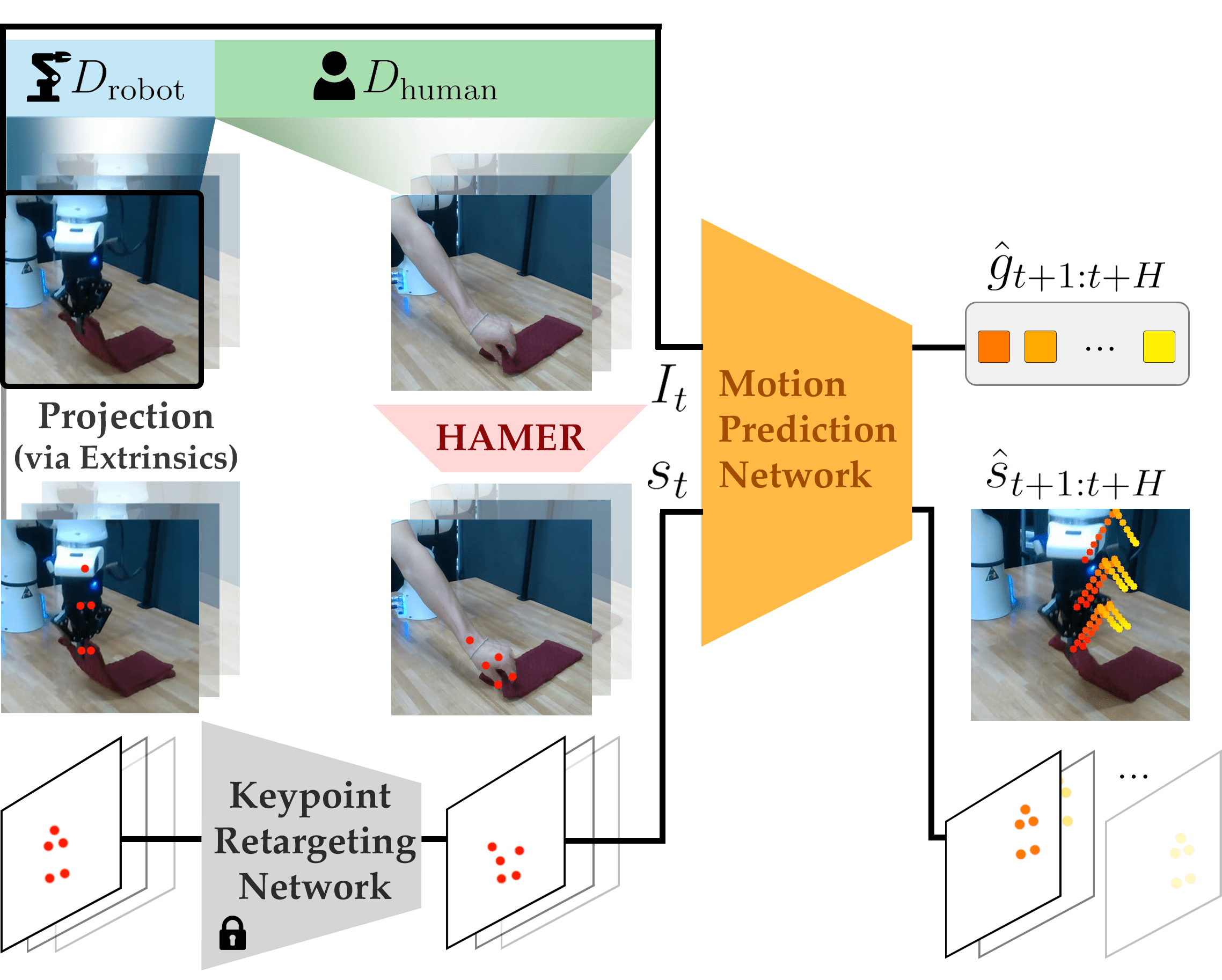}
    \caption{\textbf{MT-$\pi$ Policy Architecture.}
    We co-train MT-$\pi$ on human and robot demonstrations to predict the future pixel locations of keypoints on the end-effector (shown in red). For robot demonstrations, keypoints are extracted using calibrated camera-to-robot extrinsics, while human hand keypoints are obtained via HaMeR~\cite{pavlakos2024reconstructing}. To address embodiment differences, a Keypoint Retargeting Network maps robot keypoints to more closely resemble the human hand structure. The Motion Prediction Network, based on Diffusion Policy, takes image embeddings and current keypoints as input and predicts future keypoint tracks and grasp states. By operating entirely in image space, \mbox{MT-$\pi$} directly learns actions from both robot and human demonstrations with a cross-embodiment action representation.}
    \label{fig:mtpi_overview}
    \vspace{-0.65cm}
\end{figure}


We present \textbf{MT-}$\mathbf{\pi}$ (\textbf{M}otion \textbf{T}rack Policy), a framework designed to unify human and robot demonstrations by predicting actions for visuomotor control in image-space. Specifically, we map both human and robot demonstrations to a common representation of 2D keypoints on a manipulator in an image, and train a policy to predict the future pixel locations of these keypoints. By casting both human and robot demonstrations to this unified action space, co-training on both human and robot datasets encourages transfer between human and robot motions. At inference time, we only have to map these predictions in image space into 3D space to recover robot actions.

\subsection{Data Preprocessing}

\textbf{Robot Demonstrations.} To collect robot demonstrations, we assume access to a workspace with $\geq1$ calibrated camera (with known camera-to-robot extrinsics) and robot proprioceptive states. For each demonstration, we capture a trajectory of images $I_t^{(i)}$ from each available viewpoint. Using the robot’s end-effector position and the calibrated extrinsics, we project the 3D position of the end-effector into the 2D image plane, yielding $k$ keypoints $s_t^{(i)} = \{(u_j^{(i)}, v_j^{(i)})\}_{j=1}^k$. In practice, we take $k = 5$, giving us two points per finger on the gripper, and one in the center (\cref{fig:mtpi_overview}). We choose this positioning of points as it lends itself better to gripper positioning for grasping actions. The gripper’s open/close state is represented as a binary grasp variable $g_t^{(i)} \in \{0, 1\}$.

\textbf{Human Demonstrations.} Human demonstrations are collected using RGB cameras without needing access to calibrated extrinsics, making it possible to leverage large-scale human video datasets. We use HaMeR~\cite{pavlakos2024reconstructing}, an off-the-shelf hand pose detector, to extract a set of 21 keypoints $s_t^{(i)} = \{(u_j^{(i)}, v_j^{(i)})\}_{j=1}^{21}$.\textsuperscript{$\dagger$}\footnotetext{\textsuperscript{$\dagger$}While HaMeR does predict a 3D pose of the hand, it is trained to minimized the 2D projection error. There can be multiple poses to generate very similar projections, and thus 3D pose predictions tend to be noisy in the viewing direction of the camera~\cite{pavlakos2024reconstructing}.} To roughly match the structure of the robot gripper, we select a subset of $k = 5$ keypoints: one on the wrist and two each on the thumb and index finger. 

To infer per-timestep grasp actions from human videos, we use a heuristic based on the proximity of hand keypoints to the object(s) being manipulated. For each task, we first obtain a pixel-wise mask of the object using GroundingDINO~\cite{liu2023grounding} and SAM-v2~\cite{ravi2024sam}. Then, if the number of pixels between the object mask and the keypoints on the thumb plus any one of the other fingertips falls below some threshold, we set $g_t^{(i)} = 1$. By loosely matching the positioning and ordering of keypoints between the human hand and robot gripper, we create an explicit correspondence between human and robot action representations in the image plane.

\subsection{Training Pipeline}

\textbf{Keypoint Retargeting Network.} Despite the explicit correspondences between points on the human hand and robot gripper, the embodiment gap (e.g. size differences between hand and gripper) still induces notable distinctness in the spacing of keypoints. Directly conditioning on these points may encourage the policy to over-index on these differences, generate distinct track predictions for each embodiment, and thus fail to produce actions captured in the human demonstrations. To address this, we introduce a Keypoint Retargeting Network (\cref{fig:mtpi_overview}) that maps the robot keypoints to positions more aligned with human-hand keypoints.

For each human demonstration, we add uniform noise to all keypoints except for an anchor point (e.g., the wrist). A small MLP is trained to map these noisy keypoints back to their original positions. Once trained, the network is frozen and used during both training and testing. Since this network is trained only to map a noised version of keypoints back to their original spacing, it is compatible with either human or robot keypoint as the input. That is, any robot's keypoints will be treated as ``noisy'' and be mapped to positions that more closely resemble those of the human hand. For human keypoints, this network acts as an identity map.

\textbf{Motion Track Network.} We use a Diffusion Policy objective~\cite{chi2023diffusion} to train our policy head that predicts motion tracks. The input to the network is a concatenation of image embeddings from a pre-trained ResNet encoder~\cite{he2016deep} and the current keypoint positions $s_t^{(i)}$ in pixel space. The network then predicts offsets to each of the 2D keypoints as well as the gripper state $g_t^{(i)}$ for each future timestep over a short horizon $H$. Importantly, the model takes a single viewpoint image at a time, but is agnostic to the viewpoint pose, making it adaptable to Internet-scale human videos that often come from a single RGB viewpoint.

\textbf{Image Representation Alignment.} To discourage the policy from over-attending to the visual differences between human and robot input images, we use two auxiliary losses to encourage alignment of their visual embeddings: 
    \begin{itemize} 
    \item $\ell_{\rm{KL}}$: KL-Divergence Loss from~\cite{wang2023mimicplay} to minimize the divergence between human and robot feature distributions
     \item $\ell_{\rm{DA}}$: Domain Adaptation loss from~\cite{ganin2015unsupervised}, which encourages the policy to produce indistinguishable human/robot embeddings by fooling a discriminator.
\end{itemize} The total training loss is a combination of the original diffusion policy objective and a weighted sum over the auxiliary losses $\ell_{\rm{aux}} = \lambda_{\rm{KL}} \ell_{\rm{KL}} + \lambda_{\rm{DA}} \ell_{\rm{DA}}$ where $\ell_{\rm{KL}}$ and $\ell_{\rm{DA}}$ are scaling factors. In practice, we use $\ell_{\rm{KL}}=1.0$, and tune $\ell_{\rm{DA}} \in [0, 1]$ depending on the proportion of human to robot demonstrations.

\subsection{Action Inference} 
\label{sec:action_inference}

\begin{figure}
    \centering
    \includegraphics[width=\columnwidth]{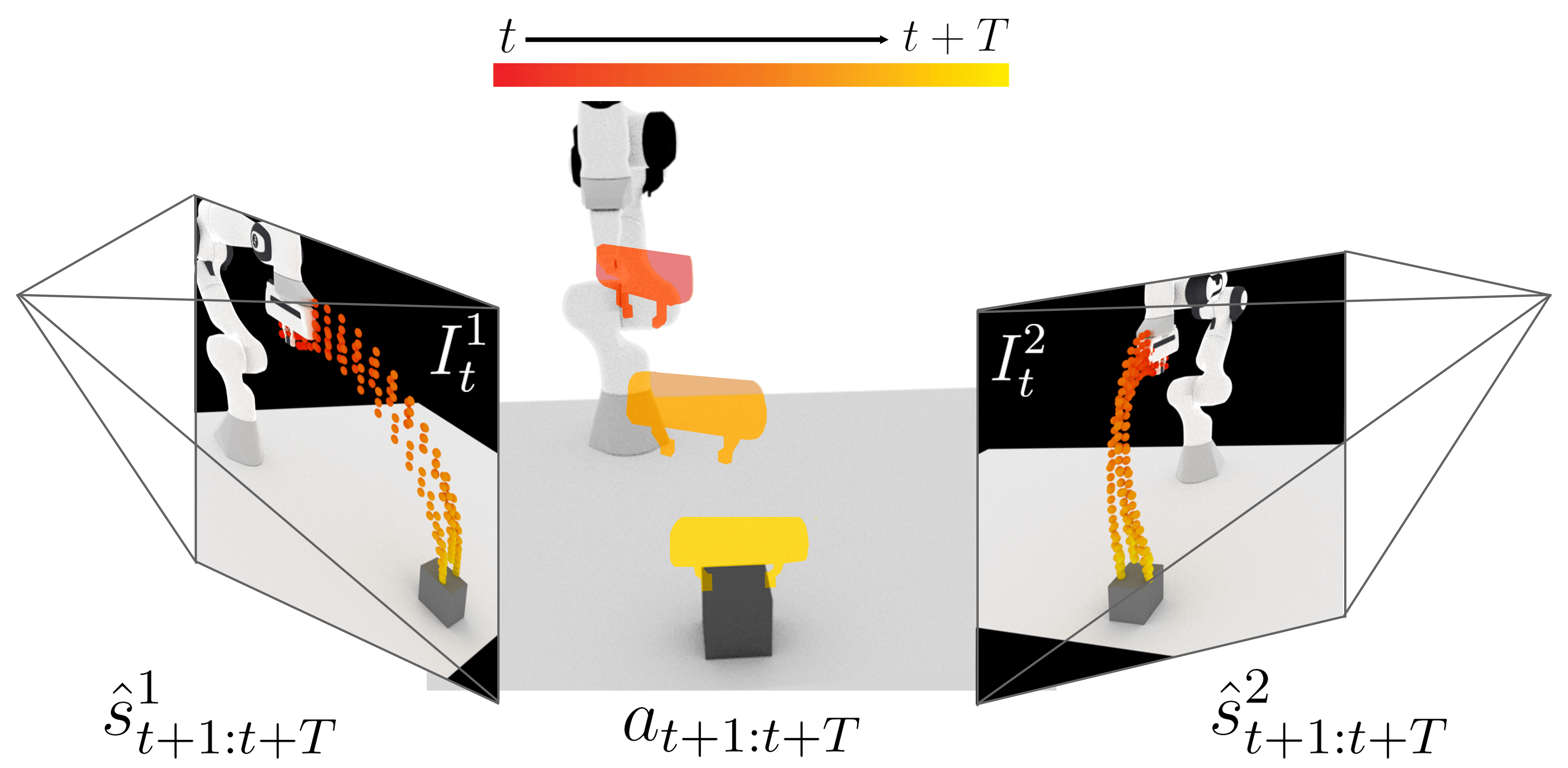}
    \caption{\textbf{MT-$\pi$ Action Inference}: MT-$\pi$ represent actions as 2D image trajectories which are not directly executable on a robot. To bridge this, we predict motion tracks from two third-person camera views and treat them as pixelwise correspondences. Using stereo triangulation with known extrinsics, we recover 3D keypoints and compute the rigid transformation between consecutive timesteps. This yields a 6DoF trajectory, $a_{t:t+T}$, for robot execution. In practice, we use a much shorter prediction horizon ($H=16 \ll T$) for more closed-loop reasoning.}
    \label{fig:action_inference}
    \vspace{-0.5cm}
\end{figure}

At inference, we assume access to 2 cameras with known extrinsics. We first leverage the camera-to-robot extrinsics to get a set of keypoints $s_t^{(i)}$ of the robot gripper per viewpoint. We then pass these keypoints through the frozen keypoint retargeting network to obtain $\hat{s}_t^{(i)}$ which are positioned more similarly to what has been seen during training. Then, $\hat{s}_t^{(i)}$ is concatenated with the image embedding of $I_t^{(i)}$ and passed as input to the motion track network $\pi_\theta$ to obtain the predicted tracks in each view, along with the gripper state $\hat{g}_{t+1:t+H}$ at each step (\cref{fig:mtpi_overview}). Using stereo triangulation with known extrinsics, we recover the 3D positions of tracks at each timestep (\cref{fig:action_inference}). 

Given the current and predicted future 3D keypoints, we then compute a rigid transformation consisting of a rotation matrix $R \in SO(3)$ and translation vector $t \in \mathbb{R}^3$ that best aligns the 3D keypoints between consecutive timesteps. These relative transformations then serve as 6DoF end-effector delta actions $a_{t:t+H}$ that can directly be executed. Accurate action recovery depends on motion tracks agreeing between two camera views. We empirically find that collecting human and robot demonstrations in a fairly consistent, unimodal way helps maintain this agreement between views, leading to more reliable action recovery.

\section{Experiments}
Our goal is to evaluate 
to what extent MT-$\pi$ can benefit from leveraging human video, the impact of our \emph{motion tracks} action space, and its generalization capabilities.

\subsection{Experimental Setup}

We evaluate MT-$\pi$ on a suite of table-top tasks against two commonly used image-based IL algorithms: Diffusion Policy (DP)~\cite{chi2023diffusion} and ACT~\cite{zhao2023learning}.  All algorithms are trained from 25 teleoperated robot demonstrations. For Diffusion Policy and ACT, we use the same output space as was chosen in the original implementation, which are 6DoF end-effector delta commands. Further, we equip all baselines with observations from an additional wrist camera (which we found to improve performance). MT-$\pi$ only receives the two third-person viewpoints, but is provided with roughly 10 minutes of additional human video demonstration per task. We highlight the differences across methods below:

\begin{table}[h]
\caption{\textbf{Methods At A Glance:} MT-$\pi$ shares the diffusion backbone with DP but differs by training on cross-embodiment data and using an image-based motion-track action space, unlike the 6DoF proprioceptive action space of DP and ACT. Additionally, unlike the baselines, MT-$\pi$ does not take wrist-camera observations as input, as these are typically absent in human videos. These design choices are intended to attribute differences in policy performance to the training data distribution and action space employed by policies, rather than other factors.}
\vspace{-10pt}
\label{tab:methods_comparison}
\begin{center}
\resizebox{\columnwidth}{!}{
\begin{tabular}{c c c c c}
\toprule
Method      & \makecell{Human and \\ Robot Data} & \makecell{Wrist \\ Camera Input} & \makecell{6DoF EE Delta \\ Action Space} & \makecell{Diffusion \\ Backbone} \\
\hline
\cellcolor{orange!30}\textbf{MT-$\pi$} & \cellcolor{orange!30}\checkmark & \cellcolor{orange!30}\texttimes & \cellcolor{orange!30}\texttimes & \cellcolor{orange!30}\checkmark \\
\textbf{DP}       & \texttimes & \checkmark & \checkmark & \checkmark \\
\textbf{ACT}      & \texttimes & \checkmark & \checkmark & \texttimes \\
\bottomrule
\end{tabular}
}
\end{center}
\vspace{-10pt}
\end{table}

\subsection{Key Results and Findings}
\textbf{Does MT-$\pi$ outperform baselines which do not leverage human video in the \emph{low} robot data regime?}

\begin{figure}
    \centering
    \includegraphics[width=\columnwidth]{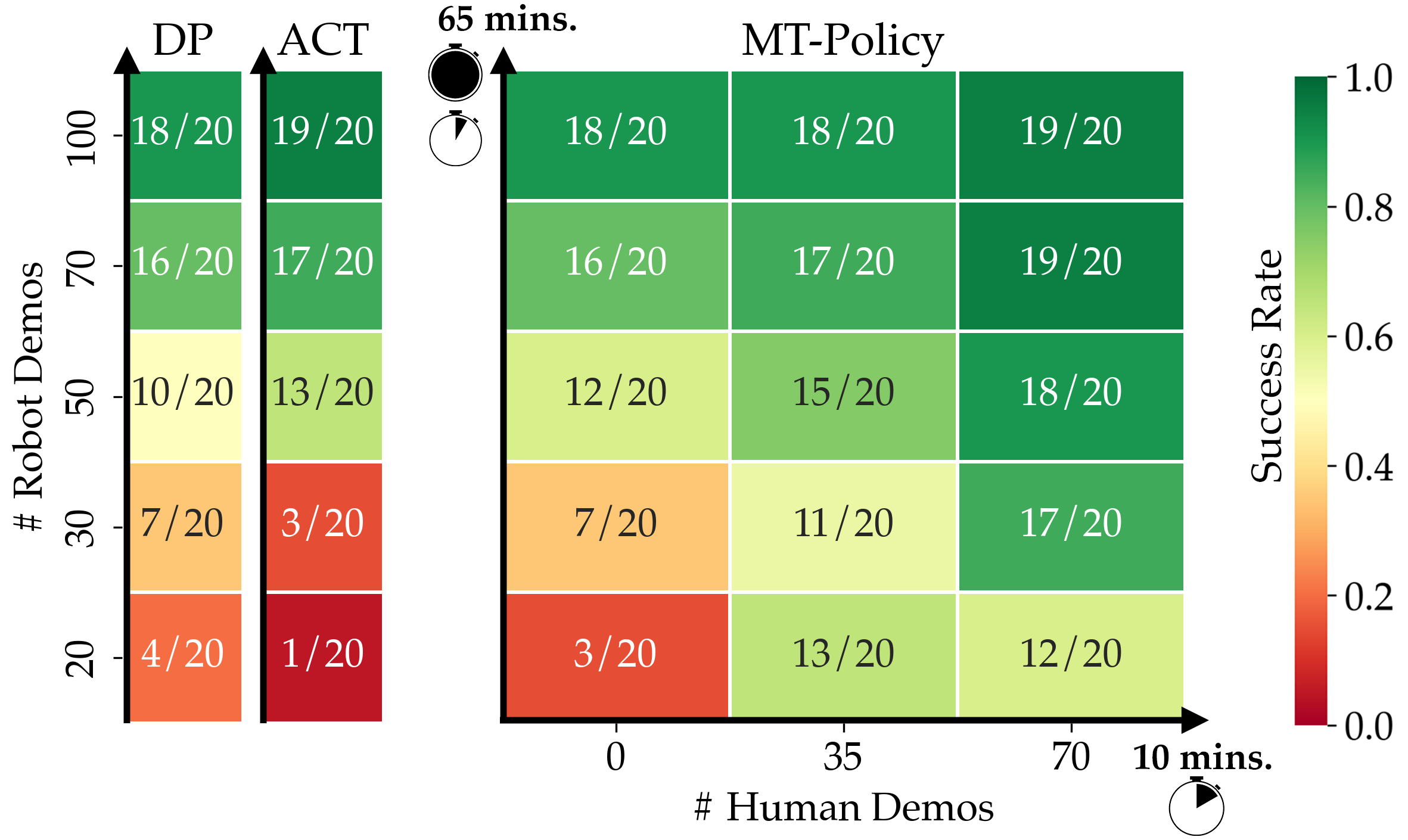}
    \caption{\textbf{Effect of Data Distribution on Policy Success:} We evaluate MT-$\pi$ against DP and ACT on the \textbf{\textit{Serve Egg}} task (see \cref{fig:front}), measuring policy success (green = high, red = low) subject to varying amounts of robot and human training data. While all policies improve with more robot data, collecting teleoperated demonstrations takes nearly \textit{5x longer} than human videos (70 human demos $\sim$ 10 mins., 100 robot demos $\sim$ 65 mins.). MT-$\pi$ achieves strong performance even in the low robot data regime by leveraging just $5-10$ minutes of human video, suggesting further gains are possible by fixing the amount of robot data and scaling human data alone.}
    \label{fig:success_heatmap}
    \vspace{-0.65cm}
\end{figure}

We first compare MT-$\pi$ to Diffusion Policy (DP)~\cite{chi2023diffusion} and Action Chunking with Transformers (ACT)~\cite{zhao2023learning} on four real-world manipulation tasks: \textit{Fork on Plate}, \textit{Fold Cloth}, \textit{Serve Egg}, and \textit{Put Away Socks} (\cref{fig:front}). As shown in \cref{tab:real_tasks}, MT-$\pi$ significantly outperforms both baselines in terms of success rate. This result suggests that MT-$\pi$'s ability to leverage human video demonstrations contributes significantly to more robust executions, especially when human videos may capture a broader diversity of motions than robot data. Indeed, due to the extremely low amounts of robot demonstrations (25 trajectories), DP and ACT have difficulty generalizing to small changes in the starting state distribution and throughout rollouts. We note that as we scale up the amount of robot demonstrations (\cref{fig:success_heatmap}), or when the task is simple enough such that the reset distribution is fully covered (\cref{tab:close_drawer}), the performances of DP and ACT reach a much higher percentage. The performance of all policies across these tasks is best understood via videos on our project \href{https://portal-cornell.github.io/motion_track_policy/}{website (https://portal-cornell.github.io/motion\_track\_policy/)}.

\begin{table}
\centering
\begin{tabular}{c|c|c|c|c}
\toprule
& \thead{Fold\\Cloth} &  \thead{Fork on\\Plate} & 
\thead{Serve\\Egg} &
\thead{Put Away\\Socks} \\
\midrule
DP~\cite{chi2023diffusion} & 7/20 & 3/20  & 7/20 & 10/20 \\
ACT~\cite{zhao2023learning} & 14/20 & 5/20 & 3/20 & 11/20 \\
\cellcolor{orange!30}MT-$\pi$ (H+R) & \cellcolor{orange!30}\textbf{18/20} & \cellcolor{orange!30}\textbf{18/20} & \cellcolor{orange!30}\textbf{17/20} & \cellcolor{orange!30} \textbf{16/20} \\
\bottomrule
\end{tabular}
\vspace{-0.5mm}
\caption{\textbf{Success Rates Across 4 Real-World Tasks:} Across four real-world tasks, we train all methods using 25 teleoperated robot demonstrations, with MT-$\pi$ receiving an additional 10 minutes of human video. Empirically, we find that even a small amount of human video enables MT-$\pi$ to outperform baselines restricted to robot data (DP and ACT). This is particularly valuable for longer-horizon tasks like \textit{Clean Up Socks} (\cref{fig:front}) where teleoperation is more time-consuming than recording human video.}
\label{tab:real_tasks}
\vspace{-0.65cm}
\end{table}

\textbf{How sample-efficient are motion tracks as an action representation when trained only on robot data, and how much additional value do human videos provide?}

Next, we study a) whether predicting actions in 2D pixel-space leads to greater sample efficiency even with just robot data, and b) to what extent MT-$\pi$ is able to benefit from the inclusion of human video demonstrations. We compare \mbox{MT-$\pi$'s} success under varying amounts of robot and human data on a medium-complexity task, \textit{Serve Egg}, in which the goal is to pick up a pan with a fried egg from a stove top and place it on a small plate roughly the same diameter as the pan. Using a success heatmap in \cref{fig:success_heatmap}, we visualize the performance of MT-$\pi$ with varying amounts of human and robot demonstrations. We compare to DP~\cite{chi2023diffusion} and ACT~\cite{zhao2023learning} which are trained on only robot demonstrations and predict 6DoF end-effector deltas.

We note first that MT-$\pi$ without any human demonstrations matches the success rates of DP and ACT given the same amount of robot demonstrations, suggesting that predicting actions in image-space is a scalable action representation even with just robot data. More interestingly, MT-$\pi$ matches the performance of baselines despite using $40\%$ less minutes of robot demonstrations by leveraging $\sim 10$ minutes of human demonstrations. The trends of this plot further suggest that even for a fixed, small amount of teleoperated robot demonstrations, MT-$\pi$ can obtain noticeably higher policy performance simply by scaling up human video alone on the order of just minutes.

\textbf{Can MT-$\pi$ generalize to motions and objects only present in human video data?}

A benefit of motion tracks as a representation is that they allow for positive transfer of motions captured in human demonstrations to an embodied agent. This is enabled by \textit{explicitly} representing human motions within our action space, instead of only implicitly (i.e. via latent embeddings). As a result, the learned policy is no longer restricted to the coverage of actions present in the robot demonstrations. To illustrate this, consider a simple task of closing a drawer, where $D_{\textrm{robot}}$ contains only demonstrations of the drawer being closed to the right, whereas $D_{\textrm{human}}$ contains demonstrations of the drawer being closed to both the left and right. As seen in {\cref{tab:close_drawer}, an appealing property of MT-$\pi$'s action space when trained with human demonstrations is the ability to generalize to closing the drawer to the left. We note that DP and ACT also achieve high performance on in-robot-domain motions, but demonstrate no success in closing the drawer to the direction only present in human videos.
\section{Limitations and Failure Modes}
While MT-$\pi$ demonstrates sample-efficiency, generalization, and reliable performance for the most part, our policy is not without failures. Namely, our policy currently makes predictions on one image at a time, and handles different viewpoints independently. This does not explicitly enforce consistency between tracks across separate views, which can lead to triangulation errors that produce imprecise actions. We try to ensure that teleoperated demonstrations are as unimodal as possible to encourage consistency in motion recovery. In the future, we can consider more explicitly enforcing viewpoint consistency via auxiliary projection/deprojection losses. Further, as a motion-centric method, our approach remains sensitive to noise in human video inputs. This sensitivity currently limits our ability to handle truly in-the-wild videos that feature drastic viewpoint shifts, egocentric motion, quick temporal changes, or occlusion of hands. While hand tracking is a fairly reliable part of our pipeline, detection of human grasps remains an open challenge. We employ a heuristic approach at present, leveraging foundation models to infer when hands and objects are in contact. Due to some imprecision in ground truth human grasp labels, our policy occasionally prematurely or imprecisely grasps objects. Nevertheless, our framework is designed with modularity in mind, allowing us to incorporate future advancements in hand perception. 



\begin{figure}[!ht]
    \centering
    \includegraphics[width=\columnwidth]{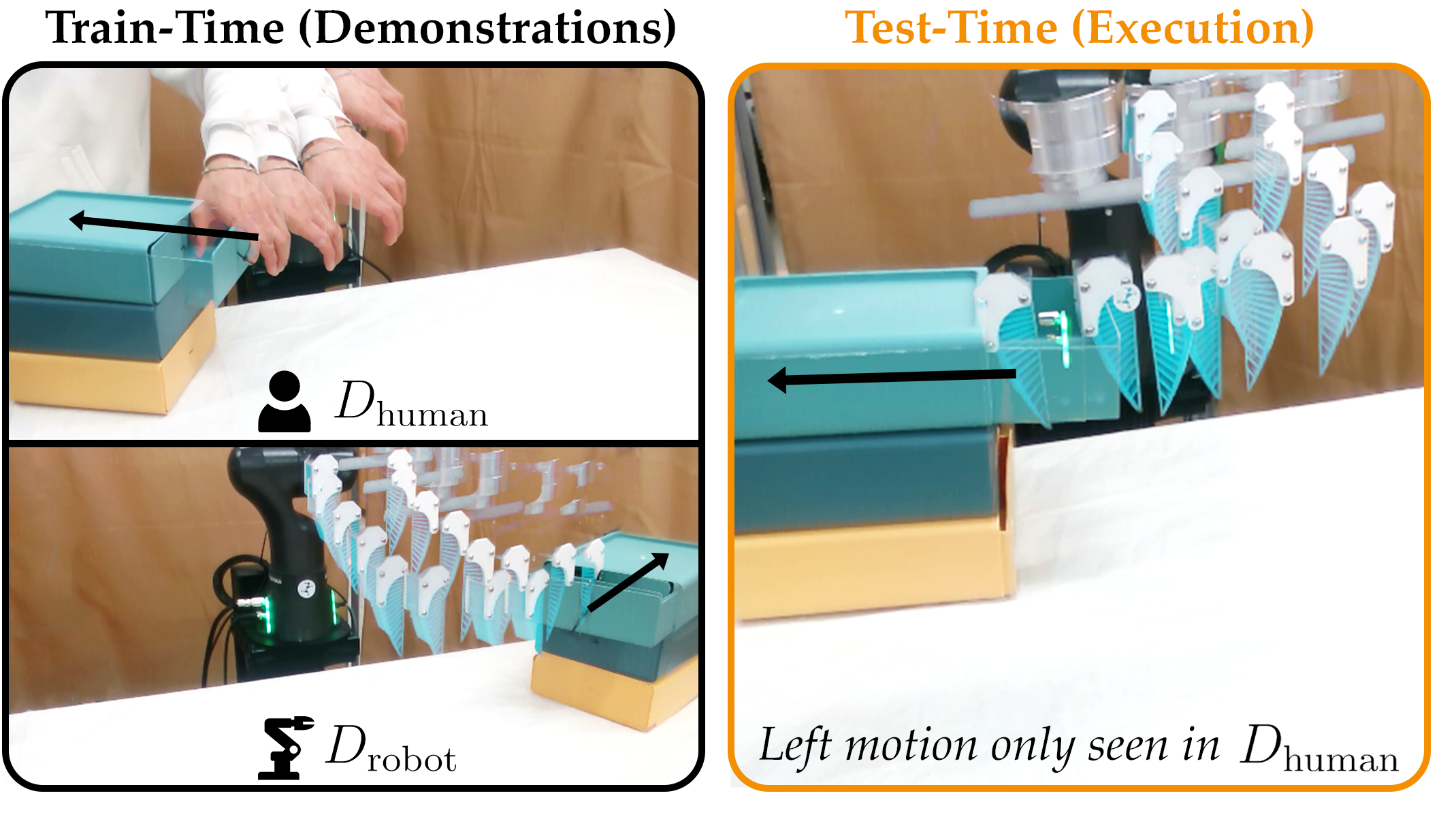}
    

    \begin{tabular}{c|c|c|c|c}
    \toprule
    \thead{\small Close Direction} & \thead{\small DP~\cite{chi2023diffusion}} & \thead{\small ACT~\cite{zhao2023learning}} & \thead{\small MT-$\pi$ \\ (Robot Only)} & \cellcolor{orange!30}\thead{\small MT-$\pi$ \\ (H+R)} \\
    \midrule
    \thead{Left \\ (In $D_{\text{robot}} \cup D_{\text{human}}$)} & \textbf{20/20} & 17/20 & \textbf{20/20} & \cellcolor{orange!30}\textbf{20/20} \\
    \thead{Right \\ (\emph{Only} in $D_{\text{human}}$)} & 0/10 & 0/10 & 0/10 & \cellcolor{orange!30}\textbf{18/20} \\
    \bottomrule
    \end{tabular}
    
    \caption{\textbf{Generalization to Motions Seen in Human-Video Only}: We evaluate two variants of MT-$\pi$ (trained on human + robot data vs. robot data only) against DP and ACT for the task of closing a drawer. Human videos include closing the drawer in both directions, while robot demonstrations only show closing to the right. While all policies perform well closing to the right (in-distribution for $D_\mathrm{robot}$), only MT-$\pi$ (H+R) generalizes to closing the drawer to the left.}
    \label{tab:close_drawer}
    \vspace{-0.65cm}
\end{figure}

\section{Discussion}
In this paper, we propose Motion Track Policy (MT-$\pi$), a novel and sample-efficient imitation learning (IL) algorithm that introduces a new cross-embodiment action space for robotic manipulation. MT-$\pi$ forecasts the future movement of manipulators via 2D trajectory predictions on images, which is feasible for both human hands and robot end-effectors. Despite this simplified representation, the approach enables full recovery of 6DoF positional and rotational robot actions via 3D reconstruction from corresponding sets of 2D tracks captured from two different views. Our empirical results show that \mbox{MT-$\pi$} outperforms state-of-the-art IL methods that omit human data or our action space on a suite of 4 real-world tasks, improving performance by 40\% on average. One of the key benefits is its compatibility with various embodiments, allowing us to rely primarily on easily accessible human video data collected in minutes, while requiring only tens of robot demonstrations for a given task. This drastically reduces the data burden typically associated with IL, while enabling generalization to novel scenarios present only in human video. In the future, we hope to extend  MT-$\pi$ to handle truly in-the-wild human videos, combine our motion-centric approach with object-centric representations obtained from foundation models, and move towards more complex manipulation tasks.

\section{Acknowledgements}

This work was supported in part by funds from the NSF Awards \#2327973, \#2006388, and \#2312956, the Office of Naval Research under ONR \#N00014-22-1-2293, the Google Faculty Research Award, as well as the OpenAI Superalignment Grant. Priya Sundaresan is supported by an NSF GRFP. We would like to thank Zi-ang Cao for their helpful feedback and suggestions. 

\bibliographystyle{./IEEEtran}
\bibliography{refs}

\end{document}